%% file: main.tex
\documentclass[letterpaper, 10 pt, conference]{ieeeconf}  

\IEEEoverridecommandlockouts                              

\usepackage[utf8]{inputenc}
\usepackage[T1]{fontenc}
\usepackage{multirow}
\usepackage{graphicx}
\usepackage{svg}
\usepackage{amsmath}
\usepackage{amsfonts,amssymb}
\usepackage{subfigure}
\usepackage[colorlinks=true, linkcolor=black]{hyperref}
\usepackage{tabularx}
\usepackage{pifont}
\newcommand{\cmark}{\ding{51}}%
\newcommand{\xmark}{\ding{55}}%
\title{\LARGE \bf
SurRoL: An Open-source Reinforcement Learning Centered and\\
dVRK Compatible Platform for Surgical Robot Learning
}

\author{Jiaqi Xu$^{1,*}$, Bin Li$^{2,*}$, Bo Lu$^{2}$, Yun-Hui Liu$^{2}$, Qi Dou$^{1}$, and Pheng-Ann Heng$^{1}$
\thanks{$^{1}$J. Q, Xu, Q. Dou, and P. A. Heng are with the Department of Computer Science and Engineering, The Chinese University of Hong Kong. 
Q. Dou and P. A. Heng are also with the T Stone Robotics Institute, CUHK.
}
\thanks{$^{2}$B. Li, B. Lu, and Y. H. Liu are with the Department of Mechanical and Automation Engineering, and T stone Robotics Institute, The Chinese University of Hong Kong.}%
\thanks{The first two authors contributed equally.}
\thanks{Corresponding author: Qi Dou (qidou@cuhk.edu.hk)}
}

\begin{document}

\maketitle
\thispagestyle{empty}
\pagestyle{empty}

\input{abstract}


\input{introduction}
\input{background}
\input{method}
\input{experiment}
\input{conclusion}




\section*{ACKNOWLEDGMENT}

This project was supported by Hong Kong Research Grants Council TRS Project No. T42-409/18-R, CUHK Shun Hing Institute of Advanced Engineering (project MMT-p5-20), and Hong Kong Multi-Scale Medical Robotics Center.


\bibliographystyle{IEEEtran}
\bibliography{ref}

\end{document}

%% file: abstract.tex
\begin{abstract}


Autonomous surgical execution relieves tedious routines and surgeon's fatigue.
Recent learning-based methods, especially reinforcement learning (RL) based methods, achieve promising performance for dexterous manipulation, which usually requires the simulation to collect data efficiently and reduce the hardware cost.
%
%
%
The existing learning-based simulation platforms for medical robots suffer from limited scenarios and simplified physical interactions, which degrades the real-world performance of learned policies.
In this work, we designed SurRoL, an RL-centered simulation platform for surgical robot learning compatible with the da Vinci Research Kit (dVRK).
The designed SurRoL integrates a user-friendly RL library for algorithm development and a real-time physics engine, which is able to support more PSM/ECM scenarios and more realistic physical interactions.
Ten learning-based surgical tasks are built in the platform, which are common in the real autonomous surgical execution.
%
%
We evaluate SurRoL using RL algorithms in simulation, provide in-depth analysis, deploy the trained policies on the real dVRK, and show that our SurRoL achieves better transferability in the real world.

\end{abstract}

%% file: introduction.tex
\section{INTRODUCTION}

Nowadays, robotic surgery systems, such as the da Vinci\textsuperscript{\textregistered} system, have been widely used in minimally invasive surgeries, including urology, gynecology, cardiothoracic, and many other procedures.
%
%
%
Recently, people have raised increasing interest in autonomous execution for surgical tasks or sub-tasks~\cite{haidegger2019autonomy}, especially with the help of the open-source da Vinci Research Toolkit (dVRK)~\cite{kazanzides2014open}, which significantly relieves tedious routines and reduces the surgeon's fatigue.
%
Nonetheless, substantial specific expertise of individual skills and a complicated development process are required to design the manually-tuned control policies~\cite{schulman2013case,osa2014online,sen2016automating}.

Learning-based methods, especially reinforcement learning (RL) based methods, provide a promising alternative to automating manual effort.
These approaches are able to develop controllers for complex skills and generalize to a broader range of tasks and environments~\cite{gu2017deep,andrychowicz2020learning}.
%
However, robot learning typically requires a large amount of labeled data and interactions with the environment~\cite{levine2018learning,lillicrap2016continuous,schulman2017proximal}, usually infeasible on real surgical robots due to the expensive time cost and the hardware wear and tear issue.
%


One intuitive choice to efficiently collect data and fast prototype for learning-based algorithms is to use the simulation, where we generate a set of labeled training data through the computer.
%
Preliminary works mitigate the limited access situation by proposing medical robot simulation platforms with robotics tasks~\cite{fontanelli2018v,munawar2019real}.
%
%
More recently, the learning-based platforms, dVRL~\cite{richter2019open} and UnityFlexML~\cite{tagliabue2020soft}, build the RL simulation environments for surgical robots on top of~\cite{fontanelli2018v} and Unity, paving the way for follow-up research on surgical manipulation~\cite{hwang2020applying} and perception~\cite{li2020super}. 
%
%
%
%

However, the existing learning-based platforms only support limited scenarios in the simulated environments~\cite{richter2019open,tagliabue2020soft}, detailed in Table~\ref{table:SimulatorComparison}.
The models trained on such platforms ignore some important scenarios, such as bimanual patient side manipulator (PSM) manipulation and endoscopic camera manipulator (ECM) control.
Moreover, the physical interactions supported by the current learning-based simulators are simplified.
For example, they consider it is successfully grasping the objects when the relative distance between the jaw tip and the object is smaller than a threshold.
The modeled trained on such simulated settings may suffer from the reality gap and fail to transfer to the real world~\cite{tagliabue2020soft}.

%
%

%
%


In this work, we build a novel surgical robotic simulation platform, \textbf{SurRoL}, which is an open-source RL-centered and dVRK compatible simulation platform for \textbf{Sur}gical \textbf{Ro}bot \textbf{L}earning.
%
%
The system design of SurRoL is shown in Fig.~\ref{fig:Overview}.
Our SurRoL is able to support more surgical operation scenarios by incorporating more single-handed/bimanual PSM(s) and ECM control tasks.
Further, the designed SurRoL with carefully modeled assets can successfully deal with more realistic physical interactions.
Code is publicly available at \url{https://github.com/med-air/SurRoL}.

Our main contributions are summarized as follows:

\begin{figure*}[tp]
  \centering
  \includegraphics[width=1.0\hsize]{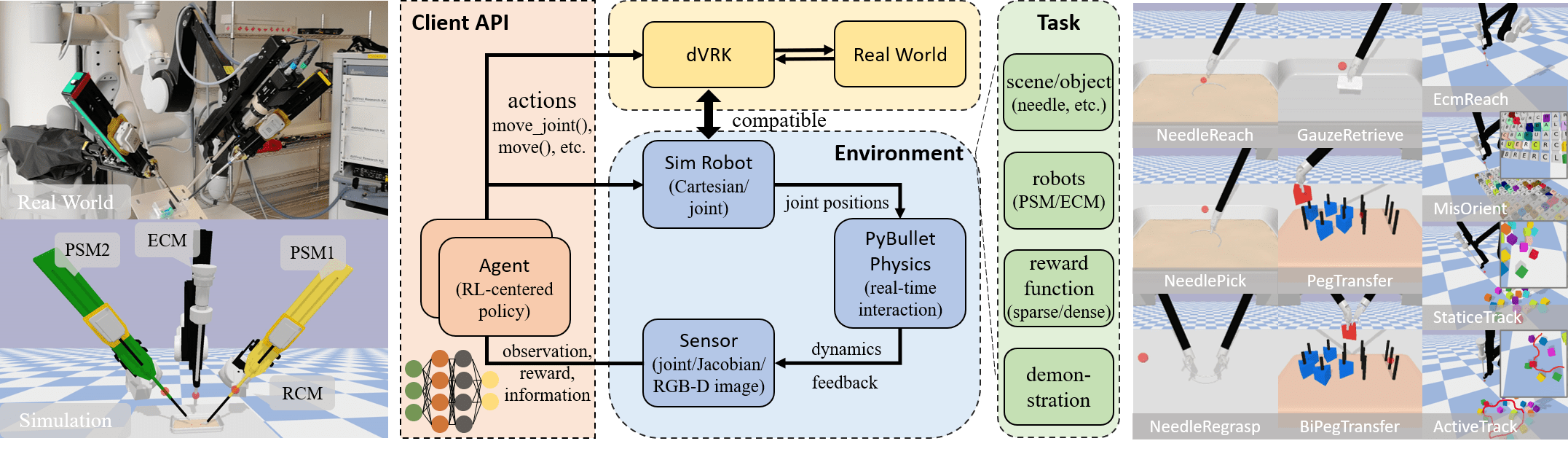}
  \caption{\textbf{System design of SurRoL.}
  SurRoL provides dVRK compatible simulation environments for surgical robot learning (left), with Gym-like interfaces for reinforcement learning algorithm development and ranges of surgical contents with physical interaction (middle).
  Ten constructed surgical relevant tasks with difficulty levels and varying scenes are presented for learning-based algorithm evaluation (right).}
  \label{fig:Overview}
\end{figure*}

\begin{table*}[tp]
    \caption{\textbf{Comparison to Existing Surgical Robot Learning Simulation Environment}}
    \centering
    \begin{tabular}{cccccccc}
    \hline
                                         & Physics & Objects             & ECM Support & Action DoF & Bimanul Task & Task Number & Interface    \\ \hline
    dVRL \cite{richter2019open}          & Static+ & Cylinder            & \xmark      & 3          & \xmark     & 2            & Python, V-REP \\
    UnityFlexML \cite{tagliabue2020soft} & Static+ & Fat tissue          & \xmark      & 3          & \xmark     & 1            & Python, Unity \\ \hline
    SurRoL (ours)                        & Dynamic & Needle, Block, etc. & \cmark      & 4          & \cmark     & 10           & Python        \\ \hline
    \multicolumn{8}{l}{Static+: grasp the object using the simplified attachment manner with limited physical interaction.}
    \end{tabular}
    \label{table:SimulatorComparison}
\end{table*}


\begin{itemize}

%
%

\item We design an open-source surgical robot learning simulation platform centered on reinforcement learning for surgical skills, which benefits low-cost data collection and accelerates the development of learning-based surgical robotic methods.
\item We build the dVRK compatible simulated environment based on the real-time physics engine, which includes diverse surgical contents and physical interaction. 
We build ten tasks (e.g., single-handed/bimanual PSM and ECM manipulation) in the platform, which are common in the real autonomous surgical execution.
\item We conduct extensive experiments for RL algorithm evaluation in simulation using the proposed tasks, provide in-depth analysis, and deploy the trained policies on the real dVRK. Results show that our SurRoL considering more rich physical interactions achieves better transferability in the real world.


\end{itemize}

%% file: background.tex
\section{RELATED WORK}

\subsection{Reinforcement Learning for Robotics}
Most of the deep RL's success for complex robotics manipulation skills originates from large amounts of interactions, using real-world robots or physics simulations.
%
Recent approaches leverage the data-driven manner to iteratively collect the data with physical robots and optimize the policy for continuous control, including grasping~\cite{levine2018learning}, poking~\cite{agrawal2016learning}, door opening~\cite{gu2017deep}, etc.
%
%
However, there are limited dVRK available worldwide with more strict safety concerns.
%
%
Alternatively, simulation is an proxy to real robots, with the benefits of low time cost and safety guarantee~\cite{ibarz2021train}.
%
%
%
Still, due to the non-trivial development process, there is no learning-based dVRK compatible simulation environment with ranges of surgical contents, tasks, and reasonable physical interaction.

\subsection{Learning Surgical Manipulation}
%
%
Previous robotics efforts in surgical skills concentrate on the sophisticated controllers specifically design for sub-tasks including, looping~\cite{osa2014online}, knot-tying~\cite{osa2014online}, needle manipulation~\cite{schulman2013case,sen2016automating}, cutting~\cite{thananjeyan2017multilateral}, tissue dissection~\cite{murali2015learning}, endoscopic guidance~\cite{osa2010framework,king2013towards}.
%
%
%
Although these carefully tuned methods can handle separate tasks reasonably, designing these algorithms exhibits substantial expertise requirements and generalization ability concerns.
%
%
%
Instead, learning-based methods, typically RL, demonstrate a significant advantage in task generalization and surgical automation with improved performance~\cite{yip2019robot}. 
Therefore, we propose an easy-to-use simulated environment with low data collection cost and state-of-the-art reinforcement learning to facilitate surgical robotics manipulation development.

\subsection{Simulation Platform for Robot Learning}
With the advancement of physics simulation and the demand for RL algorithm development, there has been a surge in robotics simulation platforms.
%
%
OpenAI Gym~\cite{1606.01540} is widely used in RL as a benchmark.
%
Other platforms focus on different features, e.g. RLBench with a wide range of manipulation tasks~\cite{james2020rlbench}, SAPIEN with home assistant robots~\cite{xiang2020sapien}, and RoboSuite with reproducible research~\cite{robosuite2020}.
%
%
%
However, surgical robots remain few attempts in simulations.
%
%
Fontanelli et al.~\cite{fontanelli2018v} relieve the situation by developing a dVRK V-REP simulator with operation scenes.
%
%
%
Though AMBF~\cite{munawar2019real} can produce dynamic environments with medical robots, it provides minimal learning environment support.
The most related works to ours are dVRL~\cite{richter2019open} and UnityFlexML~\cite{tagliabue2020soft}, reinforcement learning platforms for dVRK.
However, the low capacity of tasks with limited physical interaction restricts their functionality and sim-to-real transferability.
In this work, we develop a robot learning environment with improved scenarios and physics simulation, opening ways for future progress in surgical manipulation.

%% file: method.tex
\section{METHODS}



To provide a simulated platform for surgical robot learning, we first build a user-friendly RL library for agents to interact with.
Then, we construct the dVRK robots and surgical contents on top of the physics engine.
Finally, ten surgical learning-based tasks are built for algorithm development and evaluation.
SurRoL builds on top of the open-source PyBullet because of its state-of-the-art physics simulation, wide adoption in the machine learning community, and removal of the commercial software limits, e.g., V-REP.

\subsection{SurRoL RL Library}
\label{sebsec:RL}
SurRoL enables surgical robot learning by providing the widely used Gym-like RL environment interface for algorithm development and evaluation.

\subsubsection{Background}
Given the partially observed model of the system dynamics, we formulate the manipulation problem into the Markov Decision Process (MDP), represented by a tuple $(\mathcal{S},\mathcal{A},\mathcal{P},\mathcal{R},\gamma)$.
The agents interact with the environment, receive the current state $s_t\in \mathcal{S}$, reward $r_t\in \mathcal{R}$ based on the task specification, and generate the action $a_t=\pi(s_t)\in \mathcal{A}$ according to their policies $\pi$ at each step $t$, which forms the trajectory
$\tau=\{s_0,r_0,a_0,...,a_{t-1},s_T,r_T\}$ with a discount factor $\gamma \in [0,1)$, where $T$ denotes the episode time horizon.
The transition probability $\mathcal{P}(s_{t+1}|s_t,a_t)\in [0,1]$ is computed by the underlying physics engine.

\subsubsection{Action Space}
In practice, people frequently change the dVRK robot base frame relative to the virtual world frame, so the Cartesian-space control is used as action space, which is easier to transfer across different settings.
Though SurRoL supports six degrees of freedom (DoF) motion, we focus on the PSM tasks with the rotation within a plane in this work.
Specifically, we restrict the action space to ($d_x,d_y,d_z,d_{yaw}/d_{pitch},j$), where $d_x,d_y,d_z$ determine the position movement in the Cartesian space, $d_{yaw}/d_{pitch}$ determines the orientation movement in a top-down or vertical space setting, respectively, and $j$ determines whether the jaw is open ($j\geq0$) or close ($j<0$).
%
For ECM, besides the Cartesian-space position control, the velocity of the camera $c$ in its own fame ${}^c{V}_c$ or the roll angle control $d_{roll}$ is used when the observation is in the camera space.
%

\subsubsection{Observation Space}
%
%
SurRoL supports two observation methods, \textit{i.e.} low-dimensional ground-truth states (e.g., object 3D Cartesian positions, 6D poses), and high-dimensional RGB, depth, mask images rendered by OpenGL.
The first manner abstracts away the perception procedure and lets the agents concentrate on continuous control learning with sample efficiency.
The latter requires raw image perception, which is essential in robotic control.
In this work, we focus on the low-level continuous control skills for reinforcement learning as some built tasks are challenging even in this setting.
Unless stated otherwise, we use the low-dimensional object state (object position, orientation, etc.) and robot proprioceptive features (tip position, jaw status) represented by a fixed-length vector as the observation.

\subsubsection{Reward Function}
As reward shaping can be difficult to scale in practice \cite{ibarz2021train}, most SurRoL tasks are goal-based.
The agent receives a binary reward $r_g(s,a)=-\mathbb{I}_{f(s,a,g)}$ given the goal $g$ requirement and the condition success check function $f(s,a,g)$, and receives a negative reward unless the goal requirement is met.
While in the ECM continuous tracking task, the tracked object is constantly moving.
A dense reward function $r_d(s,a)$ is designed, which encourages the agent to follow the target.

\subsubsection{Algorithms}
Reinforcement learning algorithms aim to achieve the specified goal by learning a policy $\pi$ to maximize the expected return $\mathbb{E}_\pi[\sum_{t=0}^T\gamma^tr_t]$.
Our RL library is compatible with the popular OpenAI Gym~\cite{1606.01540}, which provides an easy-to-use interface for state-of-the-art RL algorithm evaluation and benchmark, such as DDPG~\cite{lillicrap2016continuous}, PPO~\cite{schulman2017proximal}, etc.
Meanwhile, our tasks detailed in section~\ref{subsec:tasks} involve long-horizon reasoning and can enable future research with more recent RL advances, e.g., learned skill priors~\cite{pertsch2020spirl}.

\subsection{SurRoL Physics Engine}
We build the dVRK compatible simulation environment by supporting PSM and ECM manipulation with diverse surgical contents, based on the state-of-the-art physics simulation with relatively rich robotic interactions.

\subsubsection{Physics Simulation}
We build our simulation environment based on PyBullet~\cite{coumans2016pybullet}, a Python wrapper API for the real-time Bullet physics.
Unlike previous works that approximate the grasping by attaching the object to the jaw when the tip-object relative distance is below a certain threshold \cite{richter2019open,tagliabue2020soft}, we consider more realistic scenarios by enabling inter-object physical interactions and friction-based grasping.
%
The grasping is stabilized only if the PSM can lift the grasped object above a threshold, which introduces the realism and difficulties in low-level skill learning.

\begin{figure}[tp]
    \centering
    \includegraphics[width=1.0\hsize]{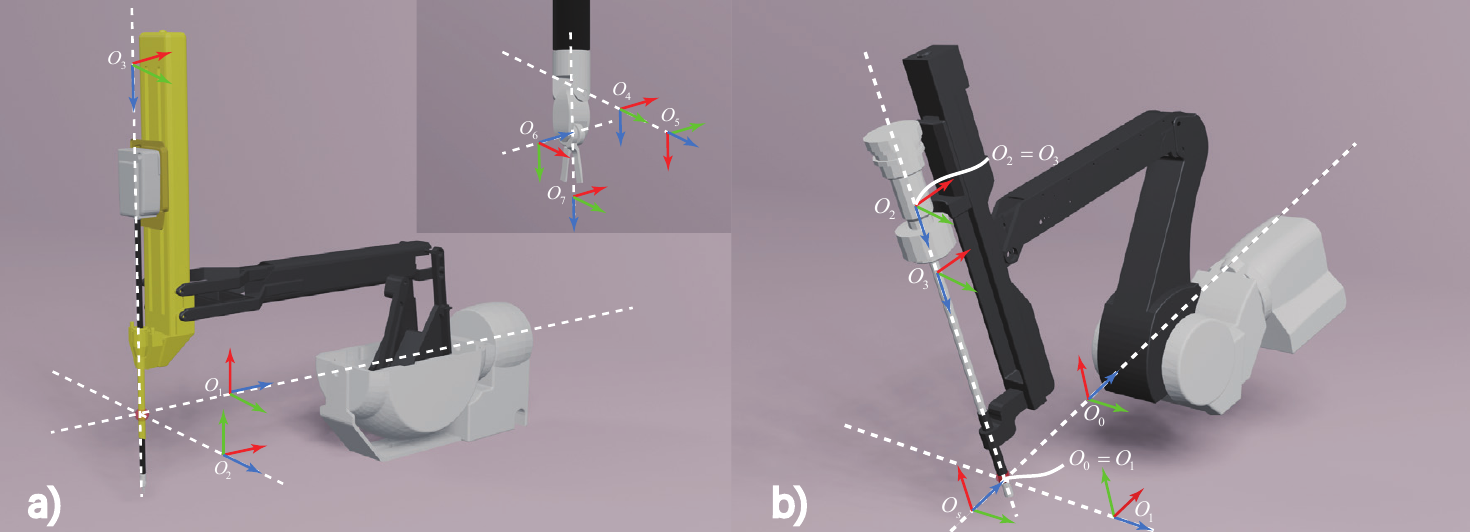}
    \caption{\textbf{PSM and ECM kinematics.}
    a) PSM is a 6-DoF actuated robot with instruments. 
    b) ECM is a 4-DoF actuated robot with the camera mounted.}
    \label{fig:PSMECM}
\end{figure}

\subsubsection{Compatible dVRK Robot}
Our simulation platform considers the manipulation of both PSM and ECM, which is compatible with the dVRK interface, as shown in Fig.~\ref{fig:PSMECM}.
We build our dVRK robots based on the meshes from AMBF~\cite{munawar2019real}.
As dVRK robots contain many redundant mechanisms with parallel linkages, we rebuild the link frames into a serially linked kinematic chain and use the built-in inverse kinematics.
%
%
While PyBullet supports the off-the-shelf velocity and torque control, the dynamics discrepancy between the simulation and the real world is more significant than position control~\cite{tan2018sim}, beyond the scope of this work.
The simulated robots behave the identical joint-space and Cartesian-space action with the real dVRK, which allows commonly used high-level control and smooth transfer.

PSM has seven DoFs, where we consider the first six DoFs ($q_i$) since the last DoF corresponds to the jaw angle.
PSM includes the revolute (R) and prismatic (P) actuated joints formed in an RRPRRR sequence (Fig.~\ref{fig:PSMECM}, a).
ECM is a 4-DoF actuated arm with an RRPR sequence (Fig.~\ref{fig:PSMECM}, b).
%
%
Note that the calculated tip pose from the forward kinematics is not the final jaw/camera pose~\cite{kazanzides2014open}.
We adopt the transformation matrix $^{tip}T_{tool}$ to transform the tip pose to the tool pose.
Finally, we acquire the tool pose $^{base}T_{tool}$ relative to the remote center of motion (RCM) as the base frame.
%

\subsubsection{Object Asset}
To enrich the manipulated contents and reflect the challenges during control, we create the SurRoL object asset (e.g., $40mm$ suture needle and pegboard), modeled using Blender.
All the articulated object links are organized in the tree structure following the URDF format.
We randomly or manually tune the object's physically related parameters, including shape, mass, friction, to mimic the real-world counterparts.
%
%
To enable reliable collision detection and physical interaction between instruments and objects, we extract the mesh convex decomposition using V-HACD~\cite{mamou2016volumetric}.




\subsection{SurRoL Task Spectrum}
\label{subsec:tasks}

\begin{table}[pt]
\caption{\textbf{SurRoL Task Spectrum Summary}}
\label{table:Task}
\begin{tabular}{ccccc}
\hline
\multicolumn{1}{c|}{}              & Arm & Action & Bimanual & Reward \\ \hline
\multicolumn{1}{c|}{NeedleReach}   & PSM & $d_{pos}$                    &          & Sparse \\
\multicolumn{1}{c|}{GauzeRetrieve} & PSM & $d_{pos},j$                  &          & Sparse \\
\multicolumn{1}{c|}{NeedlePick}    & PSM & $d_{pos},d_{yaw},j$          &          & Sparse \\
\multicolumn{1}{c|}{PegTransfer}   & PSM & $d_{pos},d_{yaw},j$          &          & Sparse \\
\multicolumn{1}{c|}{NeedleRegrasp} & PSM & $d_{pos},d_{pitch},j$        & \cmark   & Sparse \\
\multicolumn{1}{c|}{BiPegTransfer} & PSM & $d_{pos},d_{yaw},j$          & \cmark   & Sparse \\
\multicolumn{1}{c|}{EcmReach}      & ECM & $d_{pos}$                    &          & Sparse \\
\multicolumn{1}{c|}{MisOrient}     & ECM & $d_{roll}$                   &          & Sparse \\
\multicolumn{1}{c|}{StaticTrack}   & ECM & ${}^c{V}_c$                  &          & Sparse \\
\multicolumn{1}{c|}{ActiveTrack}   & ECM & ${}^c{V}_c$                  &          & Dense \\ \hline
\multicolumn{5}{l}{$d_{pos}$: $d_x,d_y,d_z$; $j$: jaw open/close.}
\end{tabular}
\end{table}

We have established a spectrum of learning-based tasks given the dexterity and precision properties in the surgical context, which covers levels of surgical skills and involves manipulating PSM(s) and ECM.
%
%
%
%
We build ten tasks with diversity, including nine goal-based tasks (four PSM single-handed tasks, two PSM bimanual tasks, three ECM tasks) and one reward-based ECM task, ranging from entry-level to sophisticated counterparts, as summarized in Table~\ref{table:Task}. 

\subsubsection{NeedleReach}
This serves as a validation task for the environment since, with hindsight experience replay~\cite{andrychowicz2017hindsight}, the policy can quickly acquire the skill.
The goal is to move the PSM jaw tip to the location slightly above the needle within a tolerance $\epsilon$, where the needle is randomly placed on a surgical tray, and the jaw is close and of fixed orientation.
%

\subsubsection{GauzeRetrieve}
Imagine that we want to retrieve the suture gauze during a surgical operation.
The goal is to sequentially pick the gauze and get it back (place it at the target position), with one DoF to indicate the jaw open/close.

\subsubsection{NeedlePick}
Based on \textit{GauzeRetrieve}, \textit{NeedlePick} involves an additional yaw angle DoF, which considers the pose of the needle.
%

\subsubsection{PegTransfer}
Peg transfer is one of the Fundamentals of Laparoscopic Surgery (FLS) tasks for hand-eye coordination~\cite{fried2004proving}, which requires collision avoidance and long-horizon reasoning.
We build a single-handed version that moves the block from one peg to the other peg without handover.

\subsubsection{NeedleRegrasp}
Initial needle grasp with one PSM often results in a non-ideal picking pose.
%
This task requires to hand over the held needle from one arm to the other arm with bimanual operations~\cite{lu2020dual}.

\subsubsection{BiPegTransfer}
This is an advanced version of \textit{PegTransfer} with bimanual operations, where the grasping arm needs to hand the block to the other arm before placing it.

\subsubsection{EcmReach}
Similar to the \textit{NeedleReach}, the goal is to move the camera mounted on ECM to a randomly sampled position.
Note that the 4th joint is fixed since it does not affect the camera position but only alters the orientation.

\subsubsection{MisOrient}
%
%
Misorientation, the difference between the camera orientation and the Natural Line-of-Sight (NLS), is inevitable during surgery since the endoscope moves under the RCM constraint.
This task requires adjusting the ECM's 4th joint such that the misorientation $\theta^{*}$ with the desired NLS is minimized, which is computed from an affine transformation $\textbf{A}$.
The goal is achieved when $\theta^{*}$ is within $\delta$.
%


\subsubsection{StaticTrack}
%
%
The goal is to let the ECM track a static target cube with red color, disturbed by other surrounding cubes, that mimics the scenario to focus on the primary instrument during surgery.
A successful tracking requires the tracked cube position $p^{ij}_t$ in image space close to the image center $p_c$ and the misorientation $\theta^{*}$ is less than $\delta$.
%

%
\begin{equation}
r_g(s,a) = - \mathbb{I}_{\left\|p^{ij}_t-p_c\right\|_2<\epsilon \cap |\theta^*|<\delta}
\label{equ:StaticTrack}
\end{equation}
%
%
%
%
%
%
%
%
%

\subsubsection{ActiveTrack}
%
%
Instead of remaining static in the given place, the target cube keeps moving and follows an online generated path at a constant speed.
The goal is to keep the ECM tracking the moving cube, with a relaxed misorientation requirement but a chance to lose the target out of the view.
%
%
A dense reward $r_d(s,a)$ is designed as follows:
%
%
%
\begin{equation}
    r_d(s,a) = C-(\left\|p^{ij}_t-p_c\right\|_2 + {\lambda \cdot |\theta^*|})
\label{equ:rew_active}
\end{equation}
where $p^{ij}_t$ and $p_c$ are the same as Equ. \ref{equ:StaticTrack}, and hyperparameters $C$ and $\lambda$ are chosen as 1 and 0.1, respectively.

\begin{figure}[tp]
  \centering
  \includegraphics[width=1.0\hsize]{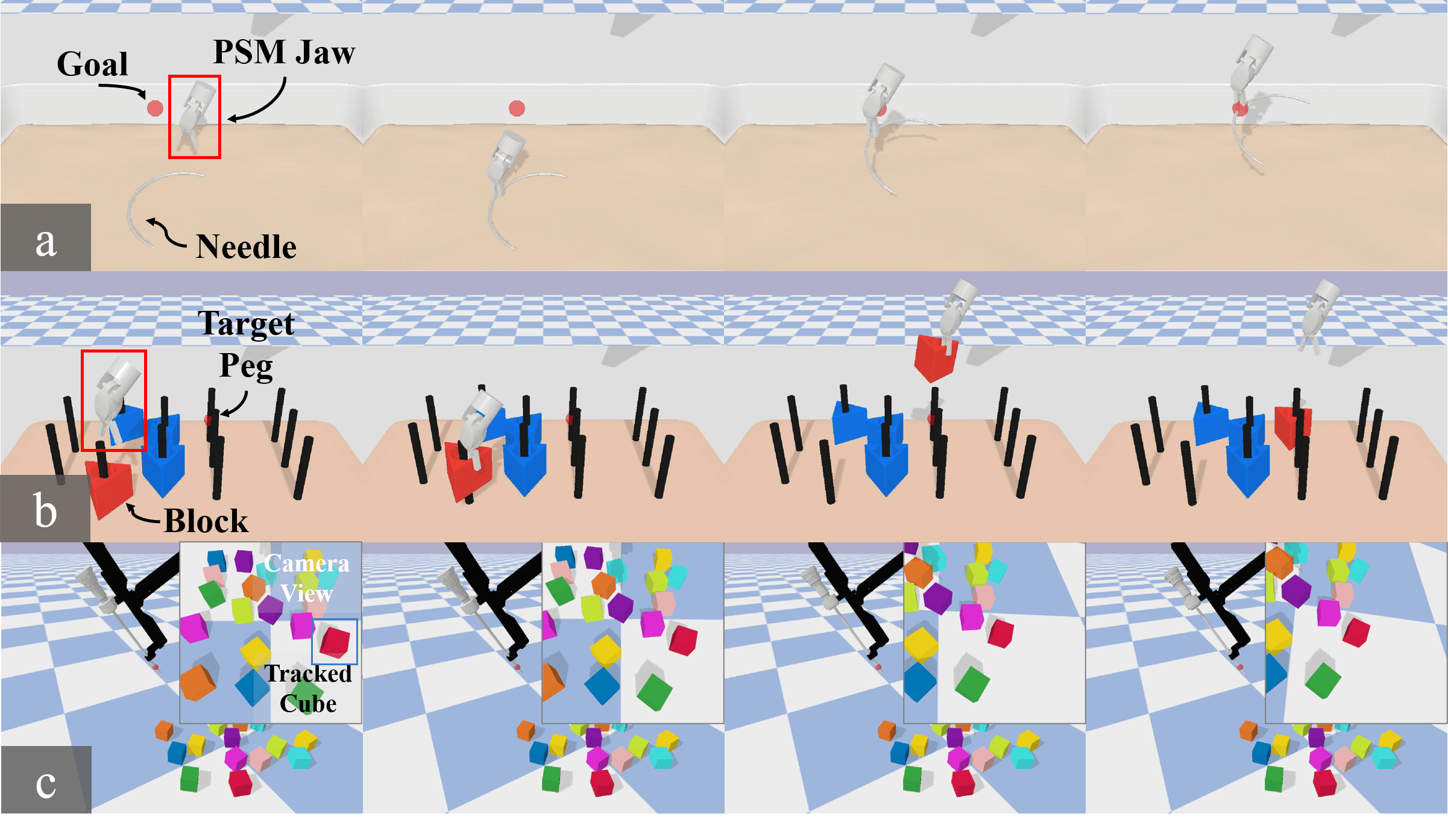}
  \caption{\textbf{Examples of the demonstration.}
  To demonstrate the proposed tasks and overcome the sample complexity, we provide the scripted policy for each and collect small amounts of demonstration data for complex ones.
  With efficient imitation learning, the policy can reasonably solve many challenging tasks that failed otherwise, e.g., \textit{PegTransfer}.}
  \label{fig:Demonstration}
\end{figure}

%% file: experiment.tex
\section{EXPERIMENTS}
%
%
In this section, we focus on the proposed learning-based tasks and want to answer the following questions:
1)~Do the tasks in our simulation platform cover a range of surgical scenarios and difficulties?
2)~Does the physical interaction matter for surgical robots with low-level skills?
3)~Can we smoothly transfer the policy trained in the simulated environment to the real dVRK?

To answer these questions, we first evaluate recent reinforcement learning algorithms optional with imitation learning in SurRoL.
%
%
Secondly, we give an in-depth analysis of the physical interaction effect using the PSM manipulation task.
Finally, we demonstrate that with the highly compatible interface, the policy from the simulation can successfully deploy on the dVRK, including PSM \textit{GauzeRetrieve}, \textit{NeedlePick}, \textit{PegTransfer}, and ECM \textit{StaticTrack}.

\subsection{Evaluation in SurRoL}
The initial experiment is to verify our proposed tasks are solvable using existing reinforcement algorithms.
As we find the manipulation tasks extremely challenging, mainly due to the tiny objects with the high precision requirement, we present the results with low-dimensional state observations.

\subsubsection{Experiment Setup}
In our RL environments, we set up the manipulation workspace for robots and objects to interact within.
For PSM tasks, the workspace is of the size $10cm^2$ and the goal tolerance distance $\epsilon =0.5cm$.
Every time the environment resets, the initial object and goal positions are randomly sampled from the workspace.
For ECM tasks, the workspace for the target cube is $80cm^2$, the misorientation tolerance $\delta=0.01rad$, and the normalized image position error $\epsilon =0.01$.
Each episode lasts for 50 timesteps for goal-based tasks and 500 timesteps for reward-based tasks.
%

For all tasks, we evaluate with the model-free RL algorithms, including the off-policy method deep deterministic policy gradient (DDPG)~\cite{lillicrap2016continuous} and the on-policy method proximal policy optimization (PPO)~\cite{schulman2017proximal}.
We collect the agent experience interacting in multiple separate environments during training and maintain a shared replay buffer for gradient update.
As model-free methods suffer from the sample complexity, we also evaluate the hindsight experience replay (HER)~\cite{andrychowicz2017hindsight}, a sample efficient learning algorithm desirable for goal-based tasks.
The success rates and episode returns are used as the evaluation metrics for goal-based and reward-based tasks, respectively, as in~\cite{andrychowicz2017hindsight,lillicrap2016continuous,schulman2017proximal}.
%

%
%

\subsubsection{Profiling Analysis}
Our SurRoL can run at a real-time rate, at about 150Hz simulation in the reaching tasks with position control and random actions, where the environment is stabilized at each time with multiple simulation steps.
Most of the training and testing experiments are performed on a desktop with Ubuntu 18.04, Inter 3.6GHz CPU with 32GB RAM, and an Nvidia TITAN RTX GPU.

\begin{figure}[t]
  \centering
  \includegraphics[width=1.0\hsize]{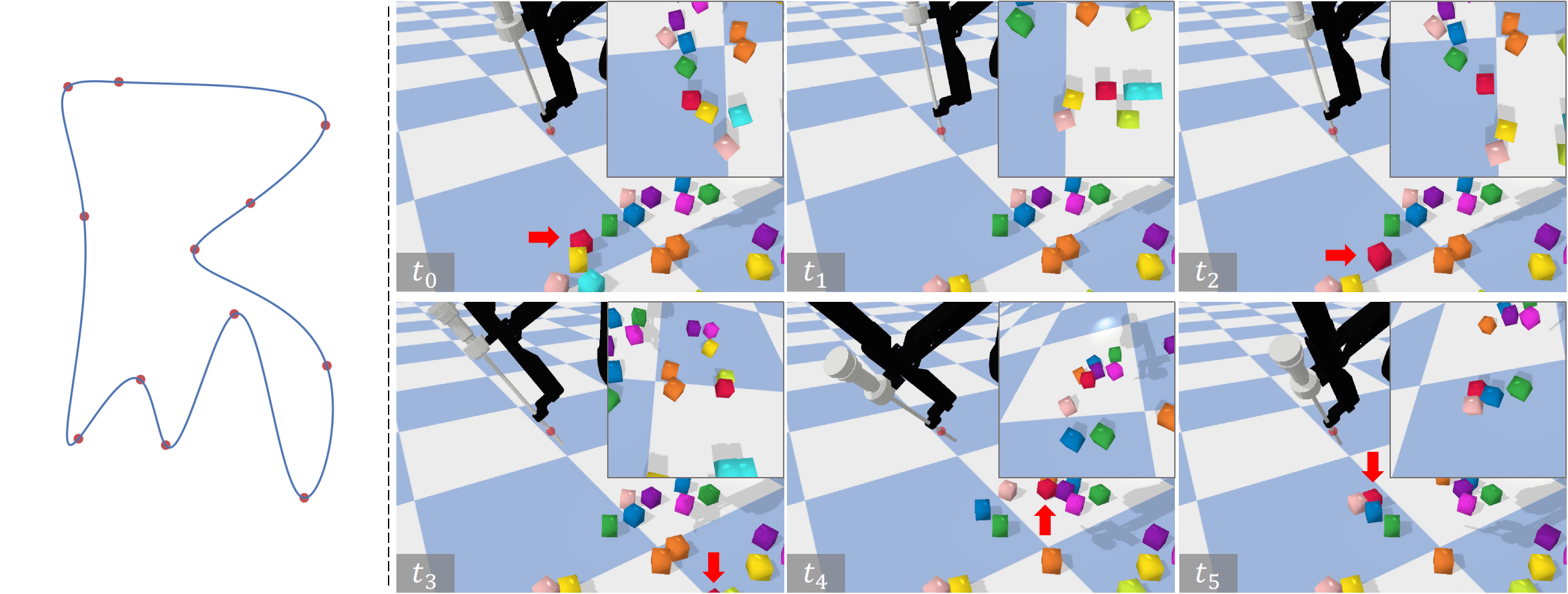}
  \caption{
  \textbf{Example of the reward-based environment ECM \textit{ActiveTrack}.}
  Each time environment resets, waypoints are sampled in the workspace randomly, generating the moving path online with B-spline interpolation (left).
  One trajectory of the policy trained using DDPG in simulation with the tracked cube marked by the red arrow is shown (right).
  }
  \label{fig:ActiveTrack}
\end{figure}

\begin{figure*}[t]
  \centering
  \includegraphics[width = 1.0\hsize]{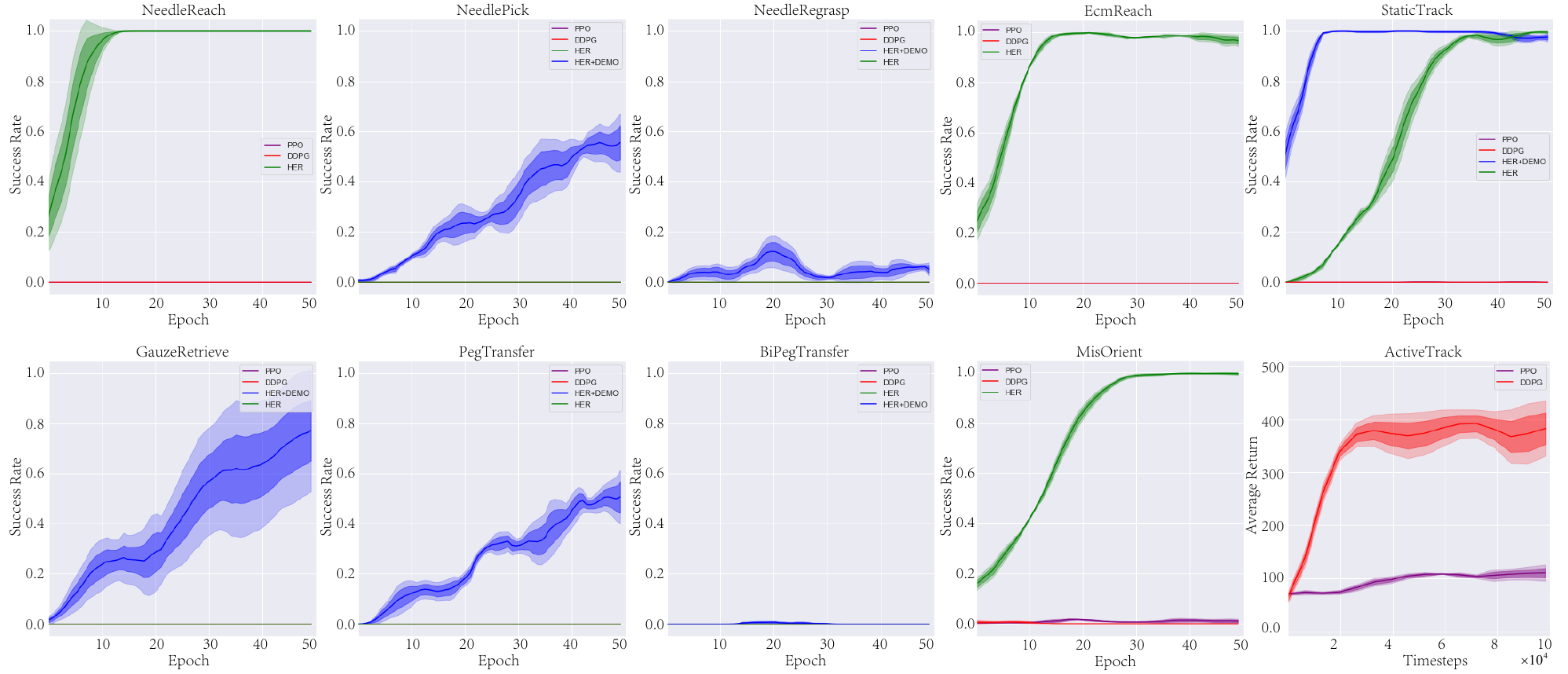}
  \caption{\textbf{Evaluation results for ten proposed tasks.} 
  The average success rates for goal-based tasks and episode returns for the reward-based task (\textit{ActiveTrack}) are shown over three random seeds, with one epoch equalling 40 episodes.
  The light shaded region denotes the standard deviation (std); the dark shaded region denotes standard error (std divided by the square root of the epoch length).}
  \label{fig:results}
\end{figure*}

\subsubsection{Demonstration using Scripted Policies}
%
%
%
To demonstrate our manipulation tasks, we design scripted policies with heuristics given the ground-truth states available in the simulation, with the help of manual engineering~\cite{ibarz2021train}.
Meanwhile, it is yet challenging to obtain satisfactory RL performance for the PSM tasks, such as \textit{NeedlePick} and \textit{PegTransfer}, which contains rich physical contacts between the instruments and the objects.
RL algorithms typically suffer from the exploration problem to discover the high reward space when the agents are trained from scratch, especially in the sparse reward setting.
To sidestep exploration challenges and ease the training, we integrate the demonstrations into the learning process by collecting a small number of samples using scripted policies for behavior cloning.

Specifically, we can divide the PSM manipulation tasks into a multi-stage sequence, where waypoints are utilized to indicate the critical changing conditions between each simplified operations.
E.g., the trajectories for \textit{NeedlePick} and \textit{PegTransfer} are composed of approaching, picking, placing, and optional releasing, as shown in Fig.~\ref{fig:Demonstration} (a), (b).
Waypoints are built manually with the position, orientation, and collision avoidance consideration, while the trajectories in-between are generated using the interpolation method.
Besides, we demonstrate the ECM tracking tasks using visual servoing, implemented by a null space method for camera velocity ${}^c{V}_c$ control as in Fig.~\ref{fig:Demonstration} (c).

\subsubsection{Evaluation Results}
%
%
A summary of the evaluation results for RL baselines is shown in Fig.~\ref{fig:results}.
For ECM goal-based tasks without instrument-object physical interaction, the agent can successfully capture the complicated action-observation relationship using HER, even for \textit{MisOrient} and \textit{StaticTrack}, which involve complex matrix transformations.
We also observe that in \textit{StaticTrack}, the learned policy can smoothly center the target object without the jittering effect, which is non-trivial for the visual servoing method that requires careful parameter tuning.
For the reward-based task, DDPG, with the proposed dense reward Equ.~\ref{equ:rew_active}, incentivizes the agent to actively control the ECM and track the moving object in a dynamic environment, which follows online generated moving paths, as illustrated in Fig.~\ref{fig:ActiveTrack}.

However, in PSM settings, HER alone cannot solve all tasks within the given time horizon, mainly due to the tiny object and physically rich interaction nature.
By visually inspecting the training progress, we find that the agents can quickly learn to approach the object such as the needle and attempt to pick reasonably, but failed because of the approximate positioning exceeding millimeters tolerance and unstable grasping.
Few experiences with high reward lead the learning to diverge in the early stage, as the policy gradually finds that random actions produce similar no-gain returns.
%

To overcome the exploration challenge, we record a small amount of demonstration data using the scripted policies for imitation learning.
After combining HER and demonstration (HER+DEMO) with Q-filtered behavior cloning~\cite{nair2018overcoming}, the agents manage to solve many challenging tasks with physics-rich simulation within 50 epochs of training, e.g. \textit{PegTransfer}.
%
%
From the results, though HER(+DEMO) performs well for robots with relatively large grippers and error tolerance~\cite{andrychowicz2017hindsight}, it performs poorly with tiny surgical instruments and objects (around 10 times smaller error tolerance), which indicates the difficulties in the medical robot field.

\begin{figure}[t]
  \centering
  \includegraphics[width=1.0\hsize]{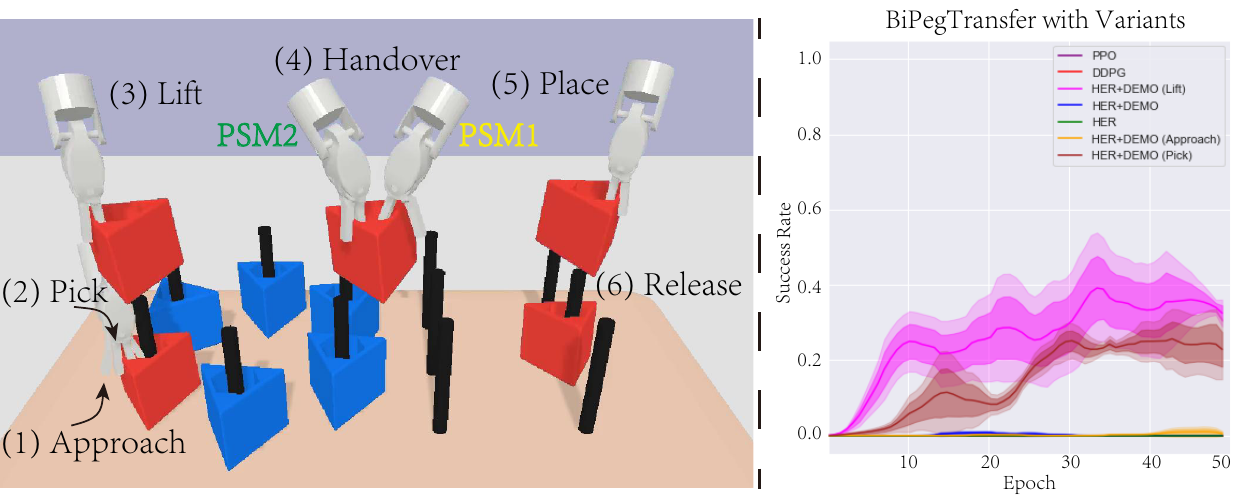}
  \caption{\textbf{Analysis of the \textit{BiPegTransfer} using HER+DEMO.}
  We analyze the difficulty for long-range skill learning by segmenting the bimanual peg transfer task into multiple steps with simplified initialization variants during environment reset (left, 1, 2, 3).
  Comparing the learning curves when the initialization is accomplished with "approach" and "pick," the stable picking skill is challenging to acquire (right).}
  \label{fig:BiPegTransfer_with_Variants}
\end{figure}

We further analyze the most challenging long-range \textit{BiPegTransfer} failed even with imitation learning by constructing several variants with different levels of simplification.
As shown in Fig.~\ref{fig:BiPegTransfer_with_Variants} left, we initialize the environment by letting the PSM2 accomplish the approach, pick and lift step manually, to inspect which part makes HER+DEMO suffer.
Surprisingly, even with the correct grasping points, HER+DEMO fails to learn the picking action, which shows the extreme exploration difficulties during learning (Fig.~\ref{fig:BiPegTransfer_with_Variants}, right).
With successful picking and lifting, the agents succeed in handing over the blocks from PSM2 to PSM1, a non-trivial coordination skill.
From the disentangled analysis, integrating motion planning and low-level control is one way to solve long-range peg transfer efficiently~\cite{hwang2020superhuman}.

\subsubsection{Physics-based Grasping Analysis}
\label{subsubsec:needle_grasp_simulation}
\begin{figure}[tp]
    \centering
    \includegraphics[width=1.0\hsize]{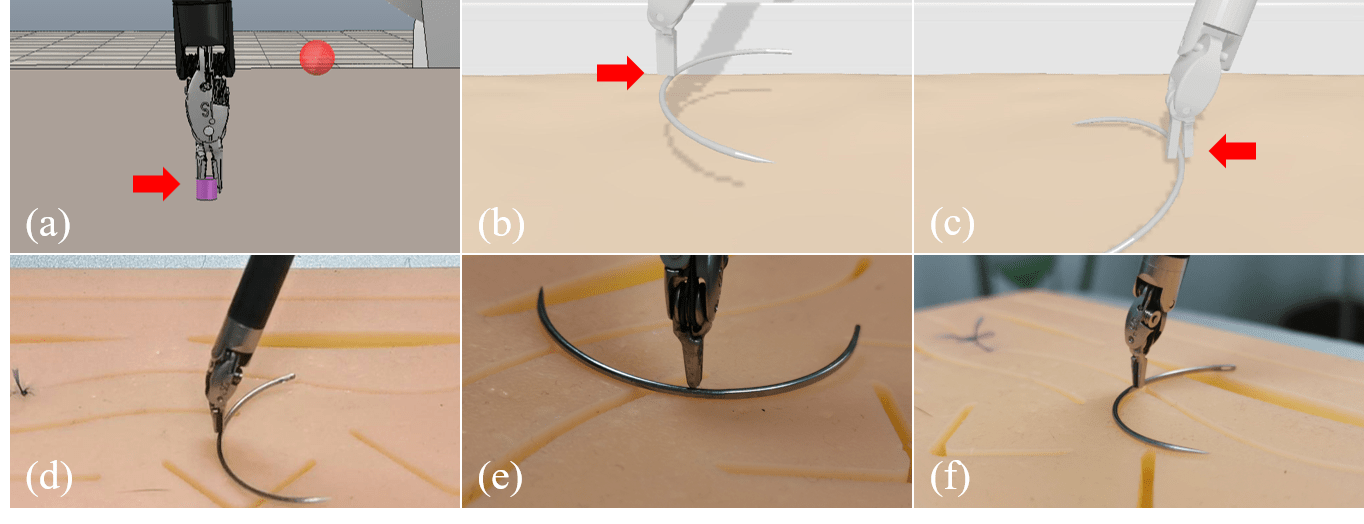}
    \caption{
    \textbf{Different levels of physical interaction.}
    The object is attached to the jaw if the tip-object distance is below a certain threshold with limited interaction~\cite{richter2019open} in (a).
    We compare the physical interaction effects by constructing a similar setting "Approx@$2mm$" with an unrealistic simulated grasp example in (b), also with our physical interaction setting "Interact" in (c).
    Some failure cases caused by the approximate picking point when deploying the trained policy using the "Approx@2$mm$" manner on the real-world dVRK from different viewpoints are shown in (d), (e), and (f).}
    \label{fig:needle_interaction}
    \vspace{-0.3cm}
\end{figure}

As we find the simplified instrument-object interaction in~\cite{richter2019open,tagliabue2020soft} may cause unstable grasping with further sim-to-real reality gap, we evaluate different physical interaction levels using \textit{NeedlePick}.
Note that the simulation backends are different among the works, so we construct a similar environment to mimic the simplified setting,
i.e., the needle is attached to the jaw when the relative distance is less than $2mm$~\cite{tagliabue2020soft}, which is denoted as "Approx@$2mm$".
Meanwhile, the needle picking point is restricted to the jaw tip to avoid unsafe jaw collisions with the holding surface.
We compare the approximate manner with ours using physical interaction and friction-based grasping (denoted as "Interact"), shown in Fig.~\ref{fig:needle_interaction} top.
The mean success rate and standard deviation of three trained policies for the two manners are presented based on the evaluation of 200 episodes per model in Table.~\ref{table:NeedleSimulation}.

We show the experimental results when the policy is trained in one physical interaction manner and tested in other settings.
Though the transition probability $\mathcal{P}$ is changed with interaction manners, polices trained with Interact are robust in the Approx settings (from 81.3\% in Interact to 70.7\% in Approx@2$mm$), which indicates the learned accurate picking points.
However, polices trained in the Approx settings suffer from dynamics confusion and significant performance degeneration while in Interact (from 76.5\% to 34.2\%) and usually fail with unrealistic no-contact grasping.
A relatively large performance improvement in a relaxed setting also reflects the inaccurately learned dynamics (from 76.5\% in Approx@$2mm$ to 88.8\% in Approx@$3mm$).

\begin{table}[tp]
    \caption{
    \textbf{The evaluation of \textit{NeedlePick} in simulation.}
    }
    \centering
    \begin{tabular}{c|cccc}
    \hline
    \multirow{2}{*}{Approach}                             & \multicolumn{4}{c}{Success Rate (\%)}                                                                                                                                           \\ \cline{2-5} 
                                                          & \begin{tabular}[c]{@{}c@{}}Approx\\ @$1mm$\end{tabular} & \begin{tabular}[c]{@{}c@{}}Approx\\ @$2mm$\end{tabular} & \begin{tabular}[c]{@{}c@{}}Approx\\ @$3mm$\end{tabular} & Interact      \\ \hline
    \begin{tabular}[c]{@{}c@{}}Approx\\ @2$mm$\end{tabular} & 36.0$\pm$12.4                                         & 76.5$\pm$4.3                                          & 88.8$\pm$5.9                                          & 34.2$\pm$16.5  \\
    Interact                                              & 52.5$\pm$9.6                                          & 70.7$\pm$19.9                                         & 76.8$\pm$14.2                                         & 81.3$\pm$1.5 \\ \hline
    \end{tabular}
    \label{table:NeedleSimulation}
    \vspace{-0.3cm}
\end{table}

\subsection{Deployment on the Real-World dVRK}
\label{subsec:deployment}
\begin{figure}[t]
  \centering
  \includegraphics[width=1.0\hsize]{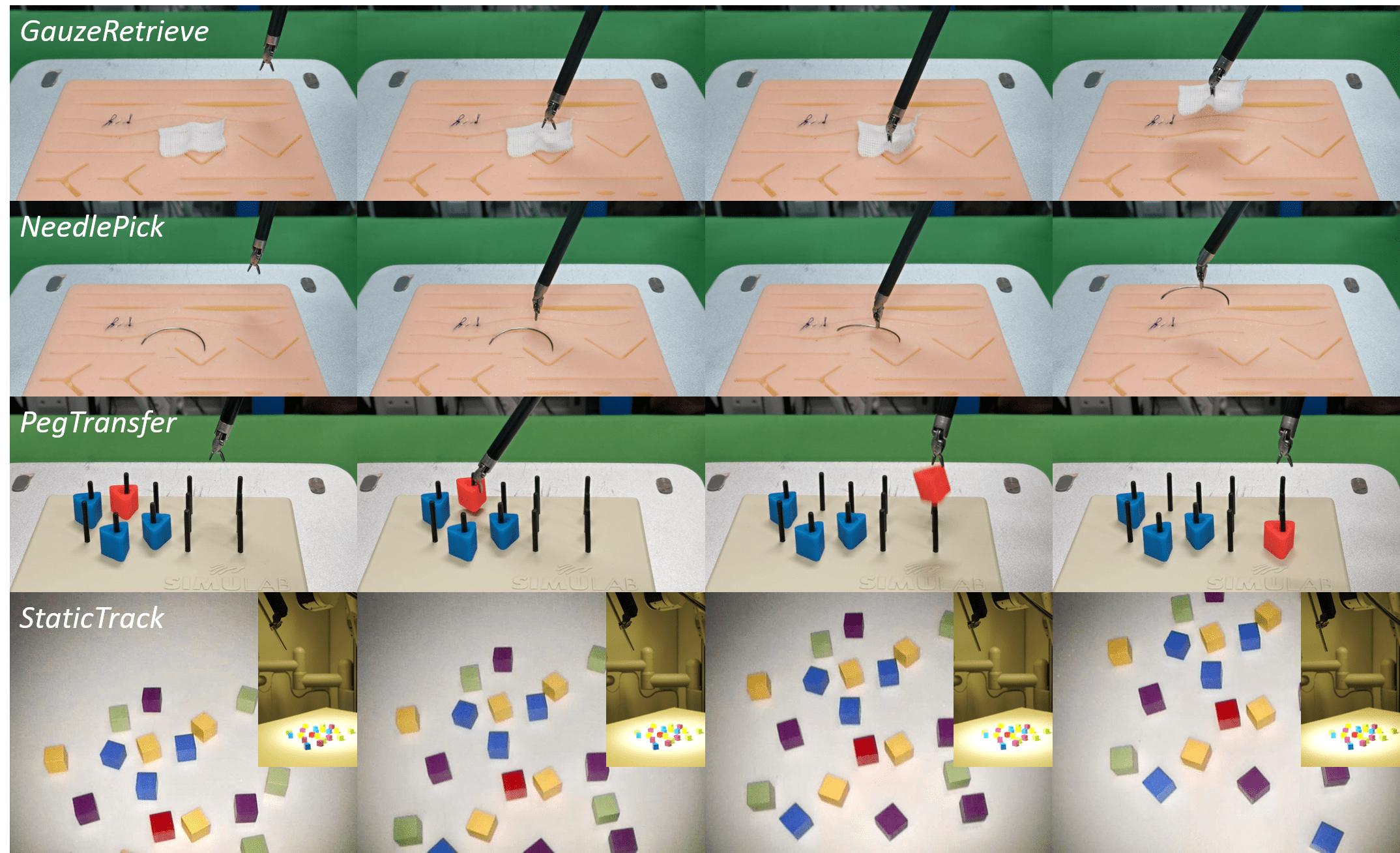}
  \caption{
  \textbf{Deployment on the real-world dVRK.}
  Trajectories of four HER(+DEMO) policies on the real dVRK demonstrate that our platform enables the smooth transfer of learned skills from the simulation to the real world. 
  More experiment details can be found in our supplementary video.
  }
  \label{fig:real_experiments}
  \vspace{-0.3cm}
\end{figure}

To demonstrate transferability, we conduct physical experiments by deploying the policies trained in SurRoL to the real-world dVRK platform.
Four tasks, PSM \textit{GauzeRetrieve}, \textit{NeedlePick}, \textit{PegTransfer}, and ECM \textit{StaticTrack}, are selected for demonstration. 
Thanks to the compatible dVRK interface, we can smoothly transfer the learned skills, with experiment snapshots shown in Fig.~\ref{fig:real_experiments}.

For the first three PSM tasks, we set up the physical experiment following the setting of~\cite{mahmood2018benchmarking} and carefully align a $10cm^2$ workspace to ensure consistency between the simulated and the real environment.
Since the tasks can be solvable only with HER+DEMO, we select the corresponding best-performance policies trained in simulation with actions generated for deployment.
With 4-DoF actions to adjust the PSM position and the jaw's open/close state, the learned \textit{GauzeRetrieve} policy can pick and retrieve the gauze to the target position within $5mm$ with a 96\% success rate on 25 episodes.
For the \textit{PegTransfer}, the learned policy sequentially picks the block, lifts it, and puts it to the target peg while avoiding collisions in the complex environment.

To investigate the reality gap that different levels of simulated interaction may cause, we conduct the physical \textit{NeedlePick} experiment using the policies introduced earlier.
We choose the best policies trained in the Approx@$2mm$ and Interact manner, with a success rate of 82.0\% and 83.5\% in their corresponding simulated settings, respectively.
The physical evaluation environments are set the same with only successful episodes for both policies in simulation to ensure fair comparisons.
The success rates are reported based on 50 episodes for each method, as shown in Table.~\ref{table:NeedleDeployment}.
From the result, the policy trained in the Approx@2$mm$ manner suffers from low real-world deployment success rates, mainly due to the imprecise picking points close but without physical contact with the needle (Fig.~\ref{fig:needle_interaction} bottom).
By contrast, the policy trained in the Interact manner with improved physics simulation is more robust to environment changes with a high success rate.
%
Besides, we find some failure cases resulting from dynamics discrepancies between the simulation and the real world, also observed in~\cite{tagliabue2020soft}.

%
%

\begin{table}[tp]
    \caption{
    \textbf{The evaluation of \textit{NeedlePick} on the real dVRK.}
    }
    \centering
    \begin{tabular}{c|cc}
    \hline
    Approach        & Trials & Success Rate (\%) \\ \hline
    Approx@2$mm$    & 29/50  & 58                \\
    Interact (ours) & 43/50  & \textbf{86}       \\ \hline
    \end{tabular}
    \label{table:NeedleDeployment}
    \vspace{-0.3cm}
\end{table}

For the ECM \textit{StaticTrack}, we mimic the simulated scene with some colored cubes, where the target cube is in red.
The best-trained policy using HER is selected to deploy into the real dVRK for ten episodes.
The target cube is segmented from the image captured by ECM first, and then the extracted position from the segmentation is served as the observation.
The policy generates joint position actions in step, converted from corresponding ${}^c{V}_{c}$ expressed in the camera frame, and center the cube in the captured image within a $0.03$ normalized position error and $0.1rad$ misorientation error, with a 90\% success rate.


%% file: conclusion.tex
\section{CONCLUSION}

In this work, we present SurRoL, a simulated platform for surgical robot learning compatible with dVRK.
Ten learning-based surgical relevant tasks with enriched assets and physical interaction are constructed, which involves manipulating PSM(s) and ECM with difficulty levels.
Extensive experiments in simulation with further physical deployment are conducted and reveal the difficulty in low-level surgical skills learning.
Moreover, the physical interaction experiments in SurRoL show that reproducing physics is one step towards a realistic simulation for surgical robot learning with transferability to the real world.
We believe SurRoL will embrace the advances in learning-based methods, especially RL and surgical robotics, to enable more researchers to be involved in the development of surgical robot manipulation.
